\documentclass[11pt,a4paper,table]{article}
\usepackage[hyperref]{naaclhlt2019}
\usepackage{times}
\usepackage{latexsym}
\usepackage{enumitem}
\usepackage{natbib}
\usepackage{graphicx}
\usepackage{colortbl}
\usepackage{xcolor}
\usepackage{url}
\usepackage{titlesec}

\aclfinalcopy 

\title{Entity Recognition at First Sight: \\
Improving NER with Eye Movement Information}

\author{Nora Hollenstein \\
  ETH Zurich\\
  {\tt noraho@ethz.ch} \\\And
  Ce Zhang \\
  ETH Zurich \\
  {\tt ce.zhang@inf.ethz.ch} \\}

\date{}

\begin{document}
\maketitle
\begin{abstract}
Previous research shows that eye-tracking data contains information about the lexical and syntactic properties of text, which can be used to improve natural language processing models. In this work, we leverage eye movement features from three corpora with recorded gaze information to augment a state-of-the-art neural model for named entity recognition (NER) with gaze embeddings. These corpora were manually annotated with named entity labels. Moreover, we show how gaze features, generalized on word type level, eliminate the need for recorded eye-tracking data at test time. The gaze-augmented models for NER using token-level and type-level features outperform the baselines. We present the benefits of eye-tracking features by evaluating the NER models on both individual datasets as well as in cross-domain settings.
\end{abstract}

\section{Introduction}

The field of natural language processing includes studies of tasks of different granularity and depths of semantics: from lower level tasks such as tokenization and part-of-speech tagging up to higher level tasks of information extraction such as named entity recognition, relation extraction, and semantic role labeling~\cite{collobert2011natural}. As NLP systems become increasingly prevalent in society, how to take advantage of information passively collected from human readers, e.g. eye movement signals, is becoming more interesting to researchers. Previous research in this area has shown promising results: Eye-tracking data has been used to improve tasks such as part-of-speech tagging \citep{barrett2016weakly}, sentiment analysis \citep{mishra2017leveraging}, prediction of multiword expressions \citep{rohanian2017using}, and word embedding evaluation \citep{sogaard2016evaluating}.

However, most of these studies focus on either relatively lower-level tasks (e.g. part-of-speech tagging and multiword expressions) or relatively global properties in the text (e.g. sentiment analysis). In this paper, we test a hypothesis on a different level: {\em Can eye movement signals also help improve higher-level semantic tasks such as extracting information from text?}

The answer to this question is not obvious. On one hand, the quality improvement attributed to eye movement signals on lower-level tasks implies that such signals do contain linguistic information. On the other hand, it is not clear whether these signals can also provide significant improvement for tasks dealing with higher-level semantics. Moreover, even if eye movement patterns contain signals related to higher-level tasks, as implied by a recent psycholinguistic study~\citep{tokunaga2017eye}, noisy as these signals are, it is not straightforward whether they would help, if not hurt, the quality of the models.

In this paper, we provide the first study of the impact of gaze features to automatic named entity recognition from text. We test the hypothesis that eye-tracking data is beneficial for entity recognition in a state-of-the-art neural named entity tagger augmented with embedding layers of gaze features. Our contributions in the current work can be summarized as follows:

\begin{enumerate}
    \item First, we manually annotate three eye-tracking corpora with named entity labels to train a neural NER system with gaze features. This collection of corpora facilitates future research in related topics. The annotations are publicly available.
    \item Beyond that, we present a neural architecture for NER, which in addition to textual information, incorporates embedding layers to encode eye movement information. 
    \item Finally, we show how gaze features generalized to word types eliminate the need for recorded eye-tracking data at test time. This makes the use of eye-tracking data in NLP applications more feasible since recorded eye-tracking data for each token in context is not required anymore at prediction time. Moreover, type-aggregated features appear to be particularly useful for cross-domain systems. 
\end{enumerate}
    
Our hypotheses are evaluated not only on the available eye-tracking corpora, but also on an external benchmark dataset, for which gaze information does not exist.

\section{Related Work}

\begin{table*}[h]
\centering
\begin{tabular}{|l|c|c|c||c|}
\hline
\textbf{} & \textbf{Dundee} & \textbf{GECO} & \textbf{ZuCo} & \textbf{Total} \\\hline
domain(s) & news articles & literature & \begin{tabular}[c]{@{}c@{}}movie reviews, \\ Wikipedia articles\end{tabular} & -  \\\hline
number of sentences & 2367 & 5424 & 700 & 8491 \\
mean sentence length & 24.75 & 12.65 & 22.12 & 19.84 \\
number of words & 58598 & 68606 & 15237 & 142441 \\
unique word types & 9131 & 5283 & 4408 & 13937\\
mean word length & 4.29 & 3.76 & 4.44 & 4.16 \\
fixation duration (ms) & 202 & 214 & 226 & 214\\
gaze duration (ms) & 237 & 232 & 265 & 244.7 \\\hline
\end{tabular}
\caption{Descriptive statistics of the eye-tracking corpora, including domain, size and mean fixation and gaze duration per token.}
\label{corpora}
\end{table*}

\begin{table*}[h]
\centering
\begin{tabular}{|l|cc|cc|cc||cc|}
\hline
\textbf{} & \multicolumn{2}{c|}{\textbf{Dundee}} & \multicolumn{2}{c|}{\textbf{GECO}} & \multicolumn{2}{c||}{\textbf{ZuCo}} & \multicolumn{2}{c|}{\textbf{Total}} \\ 
 & all & unique & all & unique & all & unique & all & unique \\ \hline
PERSON & 732 & 415 & 1870 & 108 & 657 & 446 & 3259 & 955 \\
ORGANIZATION & 475 & 261 & 26 & 12 & 156 & 95 & 657 & 364 \\ 
LOCATION & 431 & 177 & 101 & 23 & 366 & 155 & 898 & 1646 \\\hline 
\textbf{total} & \textbf{1638} & \textbf{853} & \textbf{1997} & \textbf{143} & \textbf{1179} & \textbf{696} & \textbf{4814} & \textbf{1646} \\
 &  & 52\% &  & 7\% &  & 59\% &  & 34\% \\ \hline
\end{tabular}
\caption{Number and distribution of named entity annotations in all three eye-tracking corpora.}
\label{annot-entities}
\end{table*}

The benefits of eye movement data for machine learning have been assessed in various domains, including NLP and computer vision. Eye-trackers provide millisecond-accurate records on where humans look when they are reading, and they are becoming cheaper and more easily available by the day \cite{san2009low,sewell2010real}. Although eye-tracking data is still being recorded in controlled experiment environments, this will likely change in the near future. Recent approaches have shown substantial improvements in recording gaze data while reading by using cameras of mobile devices \citep{gomez2016evaluation,papoutsaki2016webgazer}. Hence, eye-tracking data will probably be more accessible and available in much larger volumes in due time, which will facilitate the creation of sizable datasets enormously.

\citet{tokunaga2017eye} recently analyzed eye-tracking signals during the annotation of named entities to find effective features for NER. Their work proves that humans take into account a broad context to identify named entities, including predicate-argument structure. This further strengthens our intuition to use eye movement information to improve existing NER systems. And going even a step further, it opens the possibility for real-time entity annotation based on the reader's eye movements. 

The benefit of eye movement data is backed up by extensive psycholinguistic studies. For example,
when humans read a text they do not focus on every single word. The number of fixations and the fixation duration on a word depends on a number of linguistic factors \cite{clifton2007eye,demberg2008data}. First, readers are more likely to fixate on open-class words that are not predictable from context \citep{rayner1998eye}. Reading patterns are a reliable indicator of syntactical categories \citep{barrett2015reading}.  Second, word frequency and word familiarity influence how long readers look at a word. The frequency effect was first noted by \citet{rayner1977visual} and has been reported in various studies since, e.g. \citet{just1980theory} and \citet{cop2017presenting}.  Moreover, although two words may have the same frequency value, they may differ in familiarity (especially for infrequent words). Effects of word familiarity on fixation time have also been demonstrated in a number of recent studies \cite{juhasz2003investigating,williams2004eye}. Additionally, the positive effect of fixation information in various NLP tasks has recently been shown by \citet{barrett2018sequence}, where an attention mechanism is trained on fixation duration.

\paragraph{State-of-the-art NER} 
Non-linear neural networks with distributed word representations as input have become increasingly successful for any sequence labeling task in NLP \citep{huang2015bidirectional,chiu2016named,ma2016end}. The same applies to named entity recognition: State-of-the-art systems are combinations of neural networks such as LSTMs or CNNs and conditional random fields (CRFs) \citep{strauss2016results}. \citet{lample2016neural} developed such a neural architecture for NER, which we employ in this work and enhance with eye movement features. Their model successfully combines word-level and character-level embeddings, which we augment with embedding layers for eye-tracking features.

\section{Eye-tracking corpora}

For our experiments, we resort to three eye-tracking data resources: the \textit{Dundee corpus} \citep{kennedy2003dundee}, the \textit{GECO corpus} \citep{cop2017presenting} and the \textit{ZuCo corpus} \citep{hollenstein2018zuco}. For the purpose of information extraction, it is important that the readers process longer fragments of text, i.e. complete sentences instead of single words, which is the case in all three datasets.

Table \ref{corpora} shows an overview of the domain and size of these datasets. In total, they comprise 142,441 tokens with gaze information. Table \ref{corpora} also shows the differences in mean fixation times between the datasets (i.e. fixation duration (the average duration of a single fixation on a word in milliseconds) and gaze duration (the average duration of all fixations on a word)).

\paragraph{Dundee Corpus} The gaze data of the Dundee corpus \citep{kennedy2003dundee} was recorded with a \textit{Dr. Bouis Oculometer Eyetracker}. The English section of this corpus comprises 58,598 tokens in 2,367 sentences. It contains eye movement information of ten native English speakers as they read the same 20 newspaper articles from \textit{The Independent}. The text was presented to the readers on a screen five lines at a time. This data has been widely used in psycholinguistic research to analyze the reading behavior of subjects while reading sentences in context under relatively naturalistic conditions.

\paragraph{GECO Corpus} The Ghent Eye-Tracking Corpus \citep{cop2017presenting} is a more recent dataset, which was created for the analysis of eye movements of monolingual and bilingual subjects during reading. The data was recorded with an \textit{EyeLink 1000} system. The text was presented one paragraph at a time. 
The subjects read the entire novel \textit{The Mysterious Affair at Styles} by Agatha \citet{christie} containing 68,606 tokens in 5,424 sentences. We use only the monolingual data recorded from the 14 native English speakers for this work to maintain consistency across corpora.

\begin{table*}[h]
\centering
\begin{tabular}{|l|l|}
\hline
\textbf{Basic}     &      \\\hline
\textit{n} fixations     & total number of fixations on a word \textit{w}         \\
fixation probability   & the probability that a word \textit{w} will be fixated  \\
mean fixation duration & mean of all fixation durations for a word \textit{w}\\\hline
\textbf{Early}    &       \\\hline
first fixation duration & duration of the first fixation on a word \textit{w} \\
first pass duration & sum of all fixation durations during the first pass  \\\hline
\textbf{Late}  &          \\\hline
total fixation duration & sum of all fixation durations for a word \textit{w} \\
\textit{n} re-fixations   & number of times a word \textit{w} is fixated (after the first fixation)         \\
re-read probability    & the probability that a word \textit{w} will be read more than once        \\\hline
\textbf{Context}      &   \\\hline
total regression-from duration & combined duration of the regressions that began at word \textit{w} \\
\textit{w-2} fixation probability & fixation probability of the word before the previous word \\
\textit{w-1} fixation probability & fixation probability of the previous word\\
\textit{w+1} fixation probability & fixation probability of the next word \\
\textit{w+2} fixation probability & fixation probability of the word after the next word \\
\textit{w-2} fixation duration & fixation duration of the word before the previous word   \\
\textit{w-1} fixation duration & fixation duration of the previous word   \\
\textit{w+1} fixation duration & fixation duration of the next word   \\
\textit{w+2} fixation duration & fixation duration of the word after the next word  \\\hline
\end{tabular}
\caption{Gaze features extracted from the Dundee, GECO and ZuCo corpora.}
\label{feat_table}
\end{table*}

\paragraph{ZuCo Corpus} The Zurich Cognitive Language Processing Corpus \citep{hollenstein2018zuco} is a combined eye-tracking and EEG dataset. The gaze data was also recorded with an \textit{EyeLink 1000} system. The full corpus contains 1,100 English sentences read by 12 adult native speakers. The sentences were presented at the same position on the screen one at a time. For the present work, we only use the eye movement data of the first two reading tasks of this corpus (700 sentences, 15,237 tokens), since these tasks encouraged natural reading. The reading material included sentences from movie reviews from the Stanford Sentiment Treebank \citep{socher2013recursive} and the Wikipedia dataset by \citet{culotta2006integrating}. \\

For the purposes of this work, all datasets were manually annotated with named entity labels for three categories: PERSON, ORGANIZATION and LOCATION. The annotations are available at \url{https://github.com/DS3Lab/ner-at-first-sight}.

The datasets were annotated by two NLP experts. The IOB tagging scheme was used for the labeling. We followed the ACE Annotation Guidelines \cite{linguistic2005ace}. All conflicts in labelling were resolved by adjudication between both annotators. An inter-annotator reliability analysis on 10,000 tokens (511 sentences) sampled from all three datasets yielded an agreement of 83.5\% on the entity labels ($\kappa$ = 0.68). 

Table \ref{annot-entities} shows the number of annotated entities in each dataset. The distribution of entities between the corpora is highly unbalanced: Dundee and ZuCo, the datasets containing more heterogeneous texts and thus, have a higher ratio of unique entity occurrences, versus GECO, a homogeneous corpus consisting of a single novel, where the named entities are very repetitive.

\section{Eye-tracking features}\label{features}

The gaze data of all three corpora was recorded for multiple readers by conducting experiments in a controlled environment using specialized equipment. It is important to consider that, while we extract the same features for all corpora, there are certainly practical aspects that differ across the datasets. The following factors are expected to influence reading: experiment procedures; text presentation; recording hardware, software and quality; sampling rates; initial calibration and filtering, as well as \textit{human factors} such as head movements and lack of attention. Therefore, separate normalization for each dataset should better preserve the signal within each corpus and for the same reason the type-aggregation was computed on the normalized feature values. This is especially relevant for the type-aggregated features and the cross-corpus experiments described below.

In order to add gaze information to the neural network, we have selected as many features as available from those present in all three corpora. Previous research shows benefits in combining multiple eye-tracking features of different stages of the human reading process \cite{barrett2016weakly,tokunaga2017eye}. 

The features extracted follow closely on \citet{barrett2016weakly}. As described above, psycho-linguistic research has shown how fixation duration and probability differ between word classes and syntactic comprehension processes. Thus, the features focus on representing these nuances as broadly as possible, covering the complete reading time of a word at different stages. Table \ref{feat_table} shows the eye movement features incorporated into the experiments. We split the 17 features into 4 distinct groups (analogous to \citet{barrett2016weakly}), which define the different stages of the reading process:

\begin{enumerate}
    \item \textit{BASIC} eye-tracking features capture characteristics on word-level, e.g. the number of all fixations on a word or the probability that a word will be fixated (namely, the number of subjects who fixated the word divided by the total number of subjects).
    \item \textit{EARLY} gaze measures capture lexical access and early syntactic processing and are based on the first time a word is fixated.
    \item \textit{LATE} measures reflect the late syntactic processing and general disambiguation. These features are significant for words which were fixated more than once.
    \item \textit{CONTEXT} features capture the gaze measures of the surrounding tokens. These features consider the fixation probability and duration up to two tokens to the left and right of the current token. Additionally, regressions starting at the current word are also considered to be meaningful for the syntactic processing of full sentences.\\
\end{enumerate}

The eye movement measurements were averaged over all native-speaking readers of each dataset to obtain more robust estimates. The small size of eye-tracking datasets often limits the potential for training data-intensive algorithms and causes overfitting in benchmark evaluation \citep{xu2015turkergaze}. It also leads to sparse samples of gaze measurements. Hence, given the limited number of observations available, we normalize the data by splitting the feature values into quantiles to avoid sparsity issues. The best results were achieved with 24 bins. 
This normalization is conducted separately for each corpus.

Moreover, special care had to be taken regarding tokenization, since the recorded eye-tracking data considers only whitespace separation. For example, the string \textit{John's} would constitute a single token for eye-tracking feature extraction, but would be split into \textit{John} and \textit{'s} for NER, with the former token holding the label PERSON and the latter no label at all. Our strategy to address this issue was to assign the same values of the gaze features of the originating token to split tokens. 

\subsection{Type aggregation}

\citet{barrett2015using} showed that type-level aggregation of gaze features results in larger improvements for part-of-speech tagging. Following their line of work, we also conducted experiments with type aggregation for NER. This implies that the eye-tracking feature values were averaged for each word type over all occurrences in the training data. For instance, the sum of the features of all \textit{n} occurrences of the token ``island'' are averaged over the number of occurrences \textit{n}. As a result, for each corpus as well as for the aggregated corpora, a lexicon of lower-cased word types with their averaged eye-tracking feature values was compiled. Thus, as input for the network, either the type-level aggregates for each individual corpus can be used or the values from the combined lexicon, which increases the number of word types with known gaze feature values.

The goal of type aggregation is twofold. First, it eliminates the requirement of eye-tracking features when applying the models at test time, since the larger the lexicon, the more tokens in the unseen data receive type-aggregated eye-tracking feature values. For those tokens not in the lexicon, we assign a placeholder for unknown feature values. Second, type-aggregated features can be used on any dataset and show that improvements can be achieved with aggregated gaze data without requiring large quantities of recorded data.

\section{Model}

The experiments in this work were executed using an enhanced version of the system presented by \citet{lample2016neural}. This hybrid approach is based on bidirectional LSTMs and conditional random fields and relies mainly on two sources of information: character-level and word-level representations.

\begin{figure*}[h]
\centering
\includegraphics[width=0.73\textwidth]{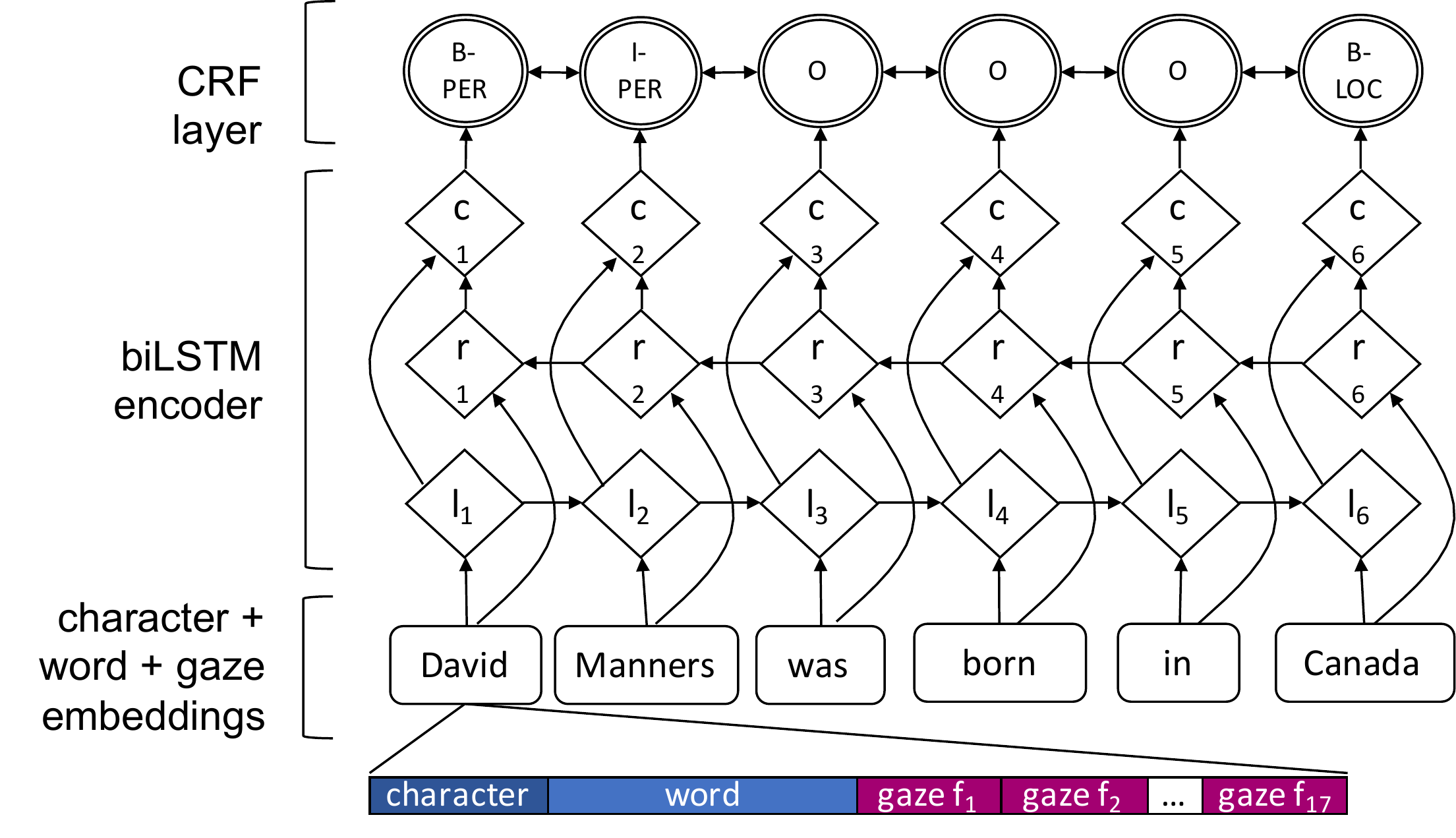}
\caption{Main architecture of the network. Character and word embeddings concatenated with gaze features are given to a bidirectional LSTM. \textit{$l_i$} represents the word \textit{i} and its left context, \textit{$r_i$} represents the word \textit{i} and its right context. Concatenating these two vectors yields a representation of the word \textit{i} in its context, \textit{$c_i$}.}
\label{architecture}
\end{figure*}

For the experiments, the originally proposed values for all parameters were maintained. Specifically, the bidirectional LSTMs for character-based embeddings are trained on the corpus at hand with dimensions set to 25. The lookup table tor the word embeddings was initialized with the pre-trained GloVe vectors of 100 dimensions \citep{pennington2014glove}. The model uses a single layer for the forward and backward LSTMs. All models were trained with a dropout rate at 0.5. Moreover, all digits were replaced with zeros.

The original model\footnote{https://github.com/glample/tagger} was modified to include the gaze features as additional embedding layers to the network. The character-level representation, i.e. the output of a bidirectional LSTM, is concatenated with the word-level representation from a word lookup table. In the augmented model with eye-tracking information, the embedding for each discrete gaze feature is also concatenated to the input. The dimension of the gaze feature embeddings is equal to the number of quantiles. This architecture is shown in Figure \ref{architecture}.
Word length and word frequency are known to correlate and interact with gaze features \citep{tomanek2010cognitive}, which is why we selected a base model that allows us to combine the eye-tracking features with word- and character-level information.

\section{Results}

Our main finding is that our models enhanced with gaze features consistently outperform the baseline. As our baseline, we trained and evaluated the original models with the neural architecture and parameters proposed by \citet{lample2016neural} on the GECO, Dundee, and ZuCo corpora and compared it to the models that were enriched with eye-tracking measures. The best improvements on F$_1$-score over the baseline models are significant under one-sided t-tests (p\textless0.05).

All models were trained with 10-fold cross validation (80\% training set, 10\% development set, 10\% test set) and early stopping was performed after 20 epochs of no improvement on the development set to reduce training time. 

First, the performance on the individual datasets is tested, together with the performance of one combined dataset consisting of all three corpora (consisting of 142,441 tokens). In addition, we evaluate the effects of the type-aggregated features using individual type lexicons for each datasets, and combining the three type lexicons of each corpus. Finally, we experiment with cross-corpus scenarios to evaluate the potential of eye-tracking features in NER for domain adaptation. Both settings were also tested on an external corpus without eye-tracking features, namely the CoNLL-2003 dataset \citep{tjong2003introduction}.

\subsection{Individual dataset evaluation}

First, we analyzed how augmenting the named entity recognition system with eye-tracking features affects the results on the individual datasets. Table \ref{results} shows the improvements achieved by adding all 17 gaze features to the neural architecture, and training models on all three corpora, and on the combined dataset containing \textit{all} sentences from the Dundee, GECO and ZuCo corpora. Noticeably, adding token-level gaze features improves the results on all datasets individually \textit{and} combined, even on the GECO corpus, which yields a high baseline due to the homogeneity of the contained named entities (see Table \ref{annot-entities}).

\begin{table}[h]
\centering
\begin{tabular}{|l|ccc|}
\hline
 & \textbf{P} & \textbf{R} & \textbf{F} \\\hline
\textbf{Dundee} & \textbf{} & \textbf{} & \textbf{} \\\hline
baseline & 79.29 & 78.56 & 78.86 \\
with gaze & 79.55 & 79.27 & 79.35 \\
type individual & \textbf{81.05} & \textbf{79.37} & \textbf{80.17}* \\
type combined & 80.27 & 79.26 & 79.67 \\\hline
\textbf{Geco} & \textbf{} & \textbf{} & \textbf{} \\\hline
baseline & 96.68 & 97.24 & 96.95 \\
with gaze & \textbf{98.08} & \textbf{97.94} & \textbf{98.01}* \\
type individual & 97.72 & 97.42 & 97.57* \\
type combined & 97.76&	97.16&	97.46*   \\\hline
\textbf{ZuCo} & \textbf{} & \textbf{} & \textbf{} \\\hline
baseline & 84.52 & 81.66 & 82.92 \\
with gaze & \textbf{86.19} & \textbf{84.28} & \textbf{85.12}* \\
type individual & 84.21 & 82.61 & 83.30 \\
type combined & 83.26 & 83.37 & 83.31 \\\hline
\textbf{All} & \textbf{} & \textbf{} & \textbf{} \\\hline
baseline & 86.92 & 86.58 & 86.72 \\
with gaze & 88.72 & 89.39 & 89.03* \\
type combined & \textbf{89.04} & \textbf{89.52} & \textbf{89.26}* \\\hline
\end{tabular}
\caption{Precision (P), recall (R) and F$_1$-score (F) for all models trained on individual datasets (best results in bold; * indicates statistically significant improvements on F$_1$-score). \textit{With gaze} are models trained on the original eye-tracking features on token-level, \textit{type individual} are the models trained on type-aggregated gaze features of this corpus only, while \textit{type combined} are the models trained with type-aggregated features computed on all datasets.}
\label{results}
\end{table}

Furthermore, Table \ref{results} also presents the results of the NER models making use of the type-aggregated features instead of token-level gaze features. There are two different experiments for these type-level features: Using the features of the word types occurring in the corpus only, or using the aggregated features of all word types in the three corpora (as describe above). As can be seen, the performance of the different gaze feature levels varies between datasets, but both the original token-level features as well as the individual and combined type-level features achieve improvements over the baselines of all datasets.

To sum up, the largest improvement with eye-tracking features is achieved when combining all corpora into one larger dataset, where an additional 4\% is gained in F$_1$-score by using type-aggregated features. Evidently, a larger mixed-domain dataset benefits from the type aggregation, while the original token-level gaze features achieve the best results on the individual datasets. Moreover, the additional gain when training on all datasets is due to the higher signal-to-noise ratio of type-aggregated features from multiple datasets.

\begin{table}[t]
\centering
\begin{tabular}{|l|ccc|}
\hline
\textbf{CoNLL-2003} & \textbf{P} & \textbf{R} & \textbf{F} \\\hline
baseline & 93.89 & 94.16 & 94.03 \\
type combined & \textbf{94.38} & \textbf{94.32} & \textbf{94.35}* \\\hline
\end{tabular}
\caption{Precision (P), recall (R) and F$_1$-score (F) for using type-aggregated gaze features on the CoNLL-2003 dataset (* marks statistically significant improvement).}
\label{conll-ind}
\end{table}

\begin{table*}[]
\centering
\begin{tabular}{|ll|ccc|ccc|ccc|}
\hline
 &  &  & \textbf{Dundee} &  &  & \textbf{GECO} &  &  & \textbf{ZuCo} &  \\
 &  & \textbf{P} & \textbf{R} & \textbf{F} & \textbf{P} & \textbf{R} & \textbf{F} & \textbf{P} & \textbf{R} & \textbf{F} \\ \hline
 & baseline & \cellcolor[HTML]{9DC3E6} & \cellcolor[HTML]{9DC3E6} & \cellcolor[HTML]{9DC3E6} & 74.20 & 70.71 & 72.40 & 75.36 & 75.62 & 75.44 \\
\textbf{Dundee} & token & \cellcolor[HTML]{9DC3E6} & \cellcolor[HTML]{9DC3E6} & \cellcolor[HTML]{9DC3E6} & 75.68 & 71.54 & 73.55* & \textbf{78.85} & 74.51 & 77.02 \\
 & type & \cellcolor[HTML]{9DC3E6} & \cellcolor[HTML]{9DC3E6} & \cellcolor[HTML]{9DC3E6} & \textbf{76.44} & \textbf{77.09} & \textbf{76.75}* & 78.33 & \textbf{76.49} & \textbf{77.35} \\\hline
 & baseline & 58.91 & 34.91 & 43.80 & \cellcolor[HTML]{9DC3E6} & \cellcolor[HTML]{9DC3E6} & \cellcolor[HTML]{9DC3E6} & 68.88 & 42.49 & 52.38 \\
\textbf{GECO} & token & \textbf{59.61} & 35.62 & \textbf{44.53} & \cellcolor[HTML]{9DC3E6} & \cellcolor[HTML]{9DC3E6} & \cellcolor[HTML]{9DC3E6} & \textbf{69.18} & \textbf{44.22} & \textbf{53.81} \\
 & type & 58.39 & \textbf{35.99} & 44.44 & \cellcolor[HTML]{9DC3E6} & \cellcolor[HTML]{9DC3E6} & \cellcolor[HTML]{9DC3E6} & 67.69 & 42.36 & 52.01 \\\hline
 & baseline & 65.85 & \textbf{54.01} & 59.34 & 83.00 & \textbf{78.11} & \textbf{80.48} & \cellcolor[HTML]{9DC3E6} & \cellcolor[HTML]{9DC3E6} & \cellcolor[HTML]{9DC3E6} \\
\textbf{ZuCo} & token & \textbf{72.62} & 50.76 & 59.70 & 82.92 & 75.35 & 78.91 & \cellcolor[HTML]{9DC3E6} & \cellcolor[HTML]{9DC3E6} & \cellcolor[HTML]{9DC3E6} \\
 & type & 69.21 & 53.05 & \textbf{59.95} & \textbf{83.68} & 74.57 & 78.85 & \cellcolor[HTML]{9DC3E6} & \cellcolor[HTML]{9DC3E6} & \cellcolor[HTML]{9DC3E6} \\\hline
\end{tabular}
\caption{Cross-corpus results: Precision (P), recall (R) and F$_1$-score (F) for all models trained on one dataset and tested on another (rows = training dataset; columns = test dataset; best results in bold; * indicates statistically significant improvements). The baseline models are trained without eye-tracking features, \textit{token} models on the original eye-tracking features, and \textit{type} are the models trained with type-aggregated features computed on all datasets.
}
\label{cross}
\end{table*}

\paragraph{Evaluation on CoNLL-2003} Going on step further, we evaluate the type-aggregated gaze features on an external corpus with no eye movement information available. The CoNLL-2003 corpus \citep{tjong2003introduction} has been widely used as a benchmark dataset for NER in different shared tasks. The English part of this corpus consists of Reuters news stories and contains 302,811 tokens in 22,137 sentences.
We use this dataset as an additional corpus without gaze information. Only the type-aggregated features (based on the combined eye-tracking corpora) are added to each word. Merely 76\% of the tokens in the CoNLL-2003 corpus also appear in the eye-tracking corpora described above and thus receive type-aggregated feature values. The rest of the tokens without aggregated gaze information available receive a placeholder for the unknown feature values.

Note that to avoid overfitting we do not train on the official train/test split of the CoNLL-2003 dataset, but perform 10-fold cross validation. Applying the same experiment setting, we train the augmented NER model with gaze features on the CoNLL-2003 data and compare it to a baseline model without any eye-tracking features. We achieve a minor, but nonetheless significant improvement (shown in Table \ref{conll-ind}), which strongly supports the generalizability effect of the type-aggregated features on unseen data.

\subsection{Cross-dataset evaluation}

In a second evaluation scenario, we test the potential of eye-tracking features for NER across corpora. The goal is to leverage eye-tracking features for domain adaptation. To show the robustness of our approach across domains, we train the models with token-level and type-level features on 100\% of corpus A and a development set of 20\% of corpus B and test on the remaining 80\% of the corpus B, alternating only the development and the test set for each fold.

Table \ref{cross} shows the results of this cross-corpus evaluation. The impact of the eye-tracking features varies between the different combinations of datasets. However, the inclusion of eye-tracking features improves the results for all combinations, except for the models trained on the ZuCo corpus and tested on the GECO corpus. Presumably, this is due to the combination of the small training data size of the ZuCo corpus and the homogeneity of the named entities in the GECO corpus.

\begin{table}[h]
\centering
\begin{tabular}{|l|ccc|}
\hline
\textbf{CoNLL-2003} & \textbf{P} & \textbf{R} & \textbf{F} \\\hline
baseline & 72.80 & 56.97 & 63.92 \\
type combined & \textbf{74.56} & \textbf{60.20} & \textbf{66.61}* \\\hline
\end{tabular}
\caption{Precision (P), recall (R) and F$_1$-score (F) for using type-aggregated gaze features trained on all three eye-tracking datasets and tested on the CoNLL-2003 dataset (* marks statistically significant improvement).}
\label{all-conll}
\end{table}

\paragraph{Evaluation on CoNLL-2003} Analogous to the individual dataset evaluation, we also test the potential of eye-tracking features in a cross-dataset scenario on an external benchmark dataset. Again, we use the CoNLL-2003 corpus for this purpose. We train a model on the Dundee, GECO \textit{and} ZuCo corpora using type-aggregated eye-tracking features and test this model on the ConLL-2003 data.
Table \ref{all-conll} shows that compared to a baseline without gaze features, the results improve by 3\% F$_1$-score. These results underpin our hypothesis of the possibility of generalizing eye-tracking features on word type level, such that no recorded gaze data is required at test time.

\begin{figure}[t]
\centering
\includegraphics[scale=0.75]{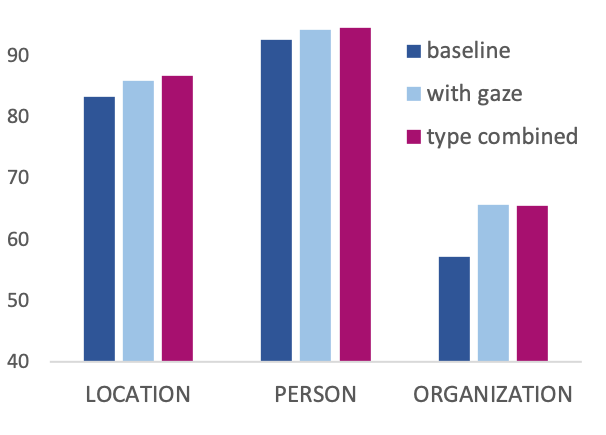}
\caption{Results per class for the models trained on all gaze datasets combined.}
\label{class}
\end{figure}

\section{Discussion}

The models evaluated in the previous section show that eye-tracking data contain valuable semantic information that can be leveraged effectively by NER systems. While the individual datasets are still limited in size, the largest improvement is observed in the models making use of \textit{all} the available data.

At a closer look, the model leveraging gaze data yield a considerably higher increase in recall when comparing to the baselines. In addition, a class-wise analysis shows that the entity type benefiting the most from the gaze features over all models is ORGANIZATION, which is the most difficult class to predict. Figure \ref{class} illustrates this with the results per class of the models trained on all three gaze corpora jointly.

In the individual dataset evaluation setting, the combined type-level feature aggregation from all datasets does not yield the best results, since each sentence in these corpora already has accurate eye-tracking features on toke-level. Thus, it is understandable that in this scenario the original gaze features and the gaze features aggregated only on the individual datasets result in better models. However, when evaluating the NER models in a cross-corpus scenario, the type-aggregated features lead to significant improvements. 

Type aggregation evidently reduces the fine-grained nuances contained in eye-tracking information and eliminates the possibility of disambiguation between homographic tokens. Nevertheless, this type of disambiguation is not crucial for named entities, which mainly consist of proper nouns and the same entities tend to appear in the same context. Especially noteworthy is the gain in the models tested on the CoNLL-2003 benchmark corpus, which shows that aggregated eye-tracking features from other datasets can be applied to any unseen sentence and show improvements, even though more than 20\% of the tokens have unknown gaze feature values. While the high number of unknown values is certainly a limitation of our approach, it shows at once the possibility of not requiring original gaze features at prediction time. Thus, the trained NER models can be applied robustly on unseen data.

\section{Conclusion}

We presented the first study of augmenting a NER system with eye-tracking information. Our results highlight the benefits of leveraging cognitive cues such as eye movements to improve entity recognition models. The manually annotated named entity labels for the three eye-tracking corpora are freely available. We augmented a neural NER architecture with gaze features. Experiments were performed using a wide range of features relevant to the human reading process and the results show significant improvements over the baseline for all corpora individually. 

In addition, the type-aggregated gaze features are effective in cross-domain settings, even on an external benchmark corpus. The results of these type-aggregated features are a step towards leveraging eye-tracking data for information extraction at training time, without requiring real-time recorded eye-tracking data at prediction time. 

\bibliography{naaclhlt2019.bib}
\bibliographystyle{acl_natbib}

\end{document}